\documentclass[lettersize,journal]{IEEEtran}
\usepackage{amsmath,amsfonts}
\usepackage{algorithmic}
\usepackage{algorithm}
\usepackage{array}
\usepackage[caption=false,font=normalsize,labelfont=sf,textfont=sf]{subfig}
\usepackage{textcomp}
\usepackage{stfloats}
\usepackage{url}
\usepackage{verbatim}
\usepackage{graphicx}
\usepackage{cite}
\usepackage{graphicx}
\usepackage{amsmath}
\usepackage{amssymb}
\usepackage{booktabs}

\usepackage{xcolor}
\usepackage{multirow}
\usepackage{mathrsfs}
\usepackage[pagebackref,breaklinks,colorlinks]{hyperref}

\begin{document}

\title{Multi-spectral Class Center Network \\ for Face Manipulation Detection and Localization}

\author{Changtao Miao, Qi Chu, Zhentao Tan, Zhenchao Jin, Tao Gong, Wanyi Zhuang, Yue Wu, Bin Liu, Honggang Hu, and Nenghai Yu
\thanks{
Email: miaoct@mail.ustc.edu.cn.
}
% \thanks{
% This work was supported by the National Natural Science Foundation of China (No. 62121002) and the Fundamental Research Funds for the Central Universities.
% Correspondence to: Qi Chu.

% Changtao Miao, Qi Chu, Tao Gong, Wanyi Zhuang, Bin Liu, Honggang Hu, and Nenghai Yu are with the School of Cyber Science and Technology, University of Science and Technology of China, Hefei 230026, China, and also with the Key Laboratory of Electromagnetic Space Information, Chinese Academy of Sciences, Hefei 230026, China (email: \{miaoct, wy970824\}@mail.ustc.edu.cn; \{qchu, tgong, flowice, hghu2005, ynh\}@ustc.edu.cn).

% Zhentao Tan and Yue Wu are with the Alibaba Cloud, Hangzhou 311121, China (email: tzt@mail.ustc.edu.cn, matthew.wy@alibaba-inc.com).

% Zhenchao Jin is with the Department of Statistics and Actuarial Science, The University of Hong Kong, Hong Kong 999077, China (email: blwx96@connect.hku.hk).

% }
}

\markboth{Journal of \LaTeX\ Class Files,~Vol.~14, No.~8, August~2021}%
{Shell \MakeLowercase{\textit{et al.}}: A Sample Article Using IEEEtran.cls for IEEE Journals}


\maketitle

\begin{abstract}
As deepfake content proliferates online, advancing face manipulation forensics has become crucial. 
To combat this emerging threat, previous methods mainly focus on studying how to distinguish authentic and manipulated face images.
Although impressive, image-level classification lacks explainability and is limited to specific application scenarios, spurring recent research on pixel-level prediction for face manipulation forensics.
However, existing forgery localization methods suffer from exploring frequency-based forgery traces in the localization network.
In this paper, we observe that multi-frequency spectrum information is effective for identifying tampered regions.
To this end, a novel \textbf{M}ulti-\textbf{S}pectral \textbf{C}lass \textbf{C}enter \textbf{Net}work (MSCCNet) is proposed for face manipulation detection and localization.
Specifically, we design a \textbf{M}ulti-\textbf{S}pectral \textbf{C}lass \textbf{C}enter (MSCC) module to learn more generalizable and multi-frequency features.
Based on the features of different frequency bands, the MSCC module collects multi-spectral class centers and computes pixel-to-class relations.
Applying multi-spectral class-level representations suppresses the semantic information of the visual concepts which is insensitive to manipulated regions of forgery images. 
Furthermore, we propose a \textbf{M}ulti-level \textbf{F}eatures \textbf{A}ggregation (MFA) module to employ more low-level forgery artifacts and structural textures.
Meanwhile, we conduct a comprehensive localization benchmark based on pixel-level FF++ and Dolos datasets.
Experimental results quantitatively and qualitatively demonstrate the effectiveness and superiority of the proposed MSCCNet.
We expect this work to inspire more studies on pixel-level face manipulation localization.
\href{https://github.com/miaoct/MSCCNet}{The codes are available.}
\end{abstract}


\section{Introduction}

%--------------Figure motivation---------------
\begin{figure}[!t]
\centering
\includegraphics[width=0.48\textwidth]{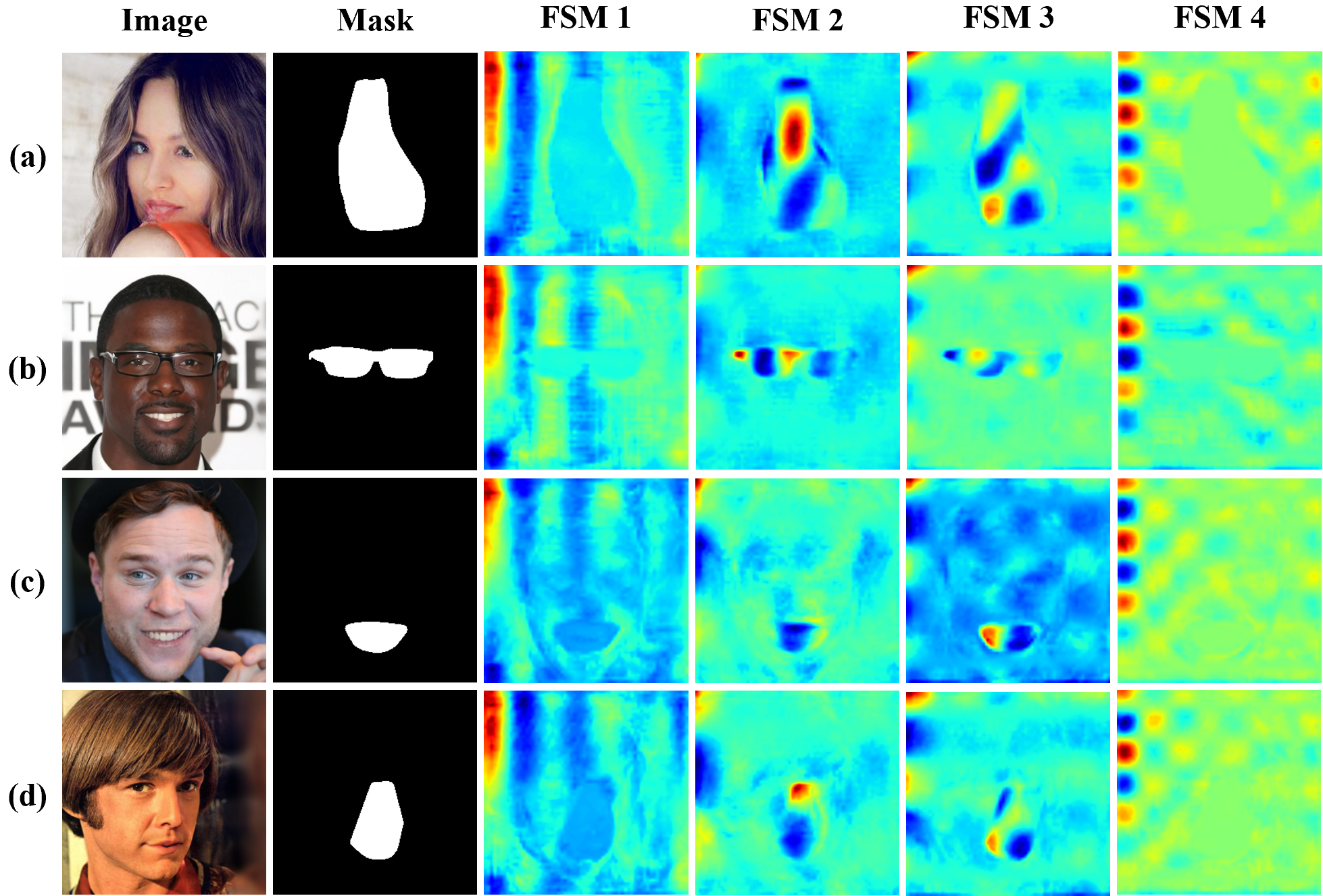}
\caption{
Visualizing the different frequency spectrum maps (FSMs) from the localization network across various manipulation methods. 
The Image column shows examples of (a) Repaint \cite{lugmayr2022RePaint}, (b) LaMa \cite{rombach2022high}, (c) LDM \cite{suvorov2021resolution}, and (d) Pluralistic \cite{zheng2019pluralistic} manipulations. The Mask column indicates the ground truth tampered regions.
Columns FSM 1-4 depict the network's multi-spectral feature maps at different frequency bands of the corresponding forged images.
Tampered areas present a more homogeneous color distribution, while authentic regions display a more heterogeneous appearance.
}
\label{fig:motivation}
\end{figure}
%--------------Figure motivation---------------

\IEEEPARstart{C}{ontinuous} advancements in Deepfake technologies \cite{deepfake2019,faceswap2019,thies2019deferred,thies2016face2face,li2019faceshifter} are resulting in the creation of remarkably realistic images and videos, exhibiting fewer noticeable tampering artifacts.
Despite their applications in the film and entertainment industries, these Deepfake tools are also exploited for malicious purposes such as creating political propaganda or pornographic content.
To address public concerns regarding misinformation, face manipulation detectors  \cite{li2018exposing,yang2019exposing,fridrich2012rich,bayar2016deep,zhao2021multi,li2018exposing,li2020sharp,qian2020thinking,luo2021generalizing,li2020face,chen2021local,wang2021m2tr, Zhao2021learning, Yu2022ImprovingGB, Shiohara2022DetectingDW} which aim to provide coarse-grained binary classification results (real or fake) at the image-level or video-level have geared extensive attentions.
The pixel-level localization of manipulated regions of Deepfake images, pivotal in analyzing and explaining the Deepfake detection results, receives inadequate attention.

Prior research \cite{dang2020detection} in face manipulation localization attempts to leverage attention maps for forged region generation, eschewing specialized localization branches. However, this approach \cite{dang2020detection} fails to capture rich global contextual information.
Subsequent methodologies \cite{nguyen2019multi, Huang2020FakeLocatorRL, Songsriin2019ComplementFF, Wu2022ExploringSF, chen2022self} widely adopt semantic segmentation pipelines, as their simple decoder networks and segmentation loss functions naturally support the face manipulation localization task. 
Nevertheless, these methods predominantly rely on vanilla RGB features, overlooking the potential of generalized frequency-based forgery cues. This oversight often leads to suboptimal generalization performance in real-world scenarios.
Recent approaches \cite{jia2021inconsistency,kong2022detect} incorporate frequency feature learning within the model's backbone network for detection tasks. However, their localization components still predominantly rely on vanilla RGB features. Concurrently, numerous face forgery detection methods  \cite{qian2020thinking, wang2021m2tr, miao2022hierarchical, miao2023f} leveraging the frequency domain demonstrate remarkable classification performance, thus validating the efficacy of frequency-based cues.
Nevertheless, these frequency modules, primarily designed for backbone networks, struggle to adapt effectively to the localization task. 
In parallel, within the image forgery localization community, several researchers \cite{Hao2021TransForensicsIF, chen2021image,kwon2022learning,guo2023hierarchical} explore the utilization of noise or frequency information to suppress image semantic object content. However, these methods typically extract noise or frequency maps directly from input RGB images, potentially limiting their capacity to capture more nuanced forgery artifacts on the feature maps.
Meanwhile, their deeper-layer features may remain semantic object information, consequently failing to preserve frequency-aware capabilities in the localization decoder network. 
Moreover, some studies \cite{Hao2021TransForensicsIF, chen2021image} demonstrate that deep semantic objective information can adversely impact the learning of tampered features.

Forgery artifacts are often accentuated in the frequency domain, with distinct characteristics manifesting across different frequency spectra. Fig. \ref{fig:motivation} illustrates this phenomenon through visualizations of various frequency band maps derived from deep features in the localization network.
For instance, as depicted in row (a) of Fig. \ref{fig:motivation}, tampered and authentic regions all exhibit markedly different patterns across the four spectral maps, demonstrating that they capture diverse forgery cues.
Motivated by these observations, we propose a novel \textbf{M}ulti-\textbf{S}pectral \textbf{C}lass \textbf{C}enter \textbf{Net}work (MSCCNet) that learns multi-spectral forgery cues to integrate frequency-domain information throughout the localization network effectively. 
The proposed method maintains frequency awareness in deeper layers while mitigating the interference of semantic information simultaneously in face manipulation localization.
The MSCCNet consists of two key components: \textbf{M}ulti-level \textbf{F}eatures \textbf{A}ggregation (MFA) and \textbf{M}ulti-\textbf{S}pectral \textbf{C}lass \textbf{C}enter (MSCC) modules. 
The proposed MFA module effectively aggregates the low-level texture information and forgery artifacts, as these cues are predominantly present in shallow features \cite{liu2021spatial,luo2021generalizing}.
The MSCC module is designed to extract the multi-frequency band forgery features and exploit class-level representations of them to suppress the semantic objective representation capability of the network.
Specifically, we first decompose the deep semantic features using a frequency transformation and calculate pixel-class relations within each spectral feature.
Then, the weighted attention of different frequency bands is acquired by computing similarity maps between different spectral class centers and the corresponding partial semantic features.
Finally, we employ weighted attention to alleviate the impact of semantic objective information and refine the original global context.
In contrast to previous methods \cite{Wu2019ManTraNetMT, Zhou2018GenerateSA, Bappy2017ExploitingSS, Hao2021TransForensicsIF, chen2021image}, our MSCC module first attempts to leverage the multi-frequency band forgery cues in the localization decoder network, and achieves satisfactory results.

The task of face manipulation localization still lacks rich pixel-level annotated datasets and standardized benchmarks, impeding the development of localization models.
To advance the face manipulation localization task, We present a reconstructed version of the FaceForensics++ (FF++) dataset \cite{rossler2019faceforensics++}, incorporating more rigorous and rational pixel-level annotations, namely P-FF++.
Subsequently, We leverage the Dolos dataset \cite{Tantaru2024WACV}, which is based on state-of-the-art GAN and diffusion models and provides official pixel-level annotations of tampered regions.
Finally, leveraging these two datasets, we construct a comprehensive benchmark to evaluate face manipulation localization models.

In a nutshell, our main contributions could be summarized as:
\begin{itemize}

   \item A novel Multi-spectral Class Center Network (MSCCNet) is designed for face manipulation localization, which consists of a Multi-level Features Aggregation (MFA) module and a Multi-spectral Class Center (MSCC) module for learning more generalizable features.

   \item We design the MSCC module to refine the original global context features by computing attention between different spectral class centers and the corresponding partial semantic features, which learns the multi-spectral forgery cues in the localization network.

   \item To facilitate the localization tasks, the reconstructed P-FF++ and diffusion-based Dolos datasets are applied to conduct a comprehensive benchmark. Extensive experiments show that our MSCCNet compares favorably against the related methods.
   
\end{itemize}

%-------------------------------------------------------------------------
\section{Related Work}\label{sec:related work}
\subsection{Face Manipulation Detection and Localization}
Early face manipulation detection methods \cite{fridrich2012rich,pan2012exposing,peng2016optimized} utilize intrinsic statistics or hand-crafted features to model spatial manipulation patterns.
Recently, data-derived detection models utilize spatial artifacts \cite{wang2021representative,song2022adaptive,song2022face,zhuang2022towards,zhuang2022uia,tan2022transformer,yang2022msta,chen2022learning,huang2023dodging,liao2023famm,li2023artifacts,wu2023interactive,hu2022detecting,zhang2024face,guo2024ldfnet} or temporal information \cite{pang2023mrenet,wang2023exploiting,yu2023msvt,yu2023pvass,wang2023spatial,zhang2024bi,liu2024mcl} to learn discriminative features and achieve remarkable detection performance. 
Others \cite{qian2020thinking,gu2021exploiting,luo2021generalizing} explore frequency information for deepfake detection, but ignore the importance of manipulated regions. 
Some studies \cite{li2020face, Zhao2021learning,chen2021local, Yu2022ImprovingGB, Shiohara2022DetectingDW,wang2021m2tr,miao2021learning,zhang2024comics} explore spatially tampered regions with segmentation loss to improve real-fake classification, but don't predict/evaluate manipulated areas. 

Recently, a few methods \cite{dang2020detection, nguyen2019multi, jia2021inconsistency,  Huang2020FakeLocatorRL, Songsriin2019ComplementFF, Wu2022ExploringSF,kong2022detect} superficially examine localization problems, but have deficiencies. 
FFD \cite{dang2020detection} applies the low-resolution attention map lacking global context. 
Some employ semantic segmentation pipelines \cite{nguyen2019multi, jia2021inconsistency,  Huang2020FakeLocatorRL, Songsriin2019ComplementFF, kong2022detect} to segment fake regions, e.g. Multi-task \cite{nguyen2019multi} designs segmentation branch. 
Prior arts \cite{Huang2020FakeLocatorRL, Songsriin2019ComplementFF} present localization for GAN-synthesized fakes, unsuitable for face manipulation data. 
Semantic segmentation networks learn semantic-dependent objects, and cannot adapt well to tampering target localization as manipulated regions are semantic-agnostic features \cite{Zhou2018GenerateSA, Bappy2017ExploitingSS}. 
We propose a multi-spectral class center module to enhance the forgery localization ability of the localization branch and suppress semantic objective information in images.

%--------------Figure framework---------------
\begin{figure*}[!t]
\centering
\includegraphics[width=0.97\textwidth]{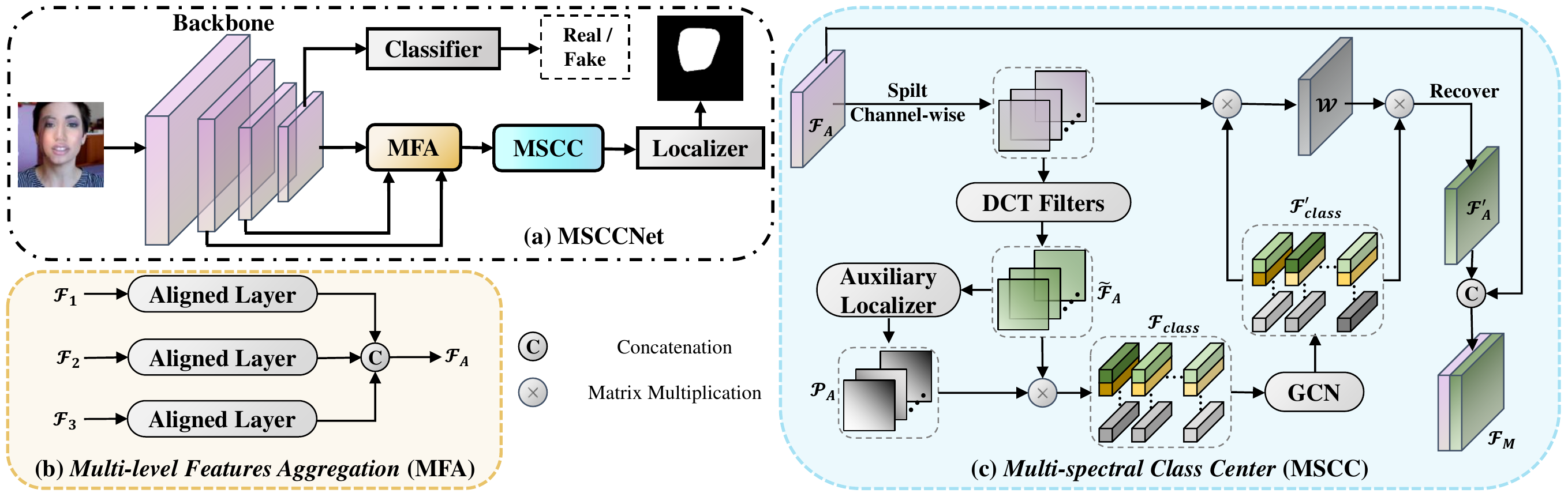}
\caption{Detailed architecture of the proposed MSCCNet. The overall network structure is shown in (a), which consists of a backbone network, a classification branch, and a localization branch. (b) shows the scheme of the forgery-related low-level texture features aggregation. (c) illustrates the process of multi-spectral class centers and different frequency attention calculations.
They are solely dedicated to enhancing the capabilities of the localization branch.
}
\label{fig:framework}
\end{figure*}
%--------------Figure framework---------------

\subsection{Image Forgery Detection and Localization}
Image forgery technologies (e.g., splicing, copy-move, removal) have existed for a long time, unlike the recent rise of face manipulation methods. 
Image forensics aims to detect spoof/bona fide images and locate tampering regions, but most image forgery localization methods \cite{li2019localization} focus only on fake datasets rather than real-fake mixed datasets. 
One localization method segments the entire input image \cite{Wu2019ManTraNetMT, Hao2021TransForensicsIF}, and another performs repeated binary classification using a sliding window \cite{Rahmouni2017DistinguishingCG}. 
% Our MSCCNet takes cropped facial areas as input, reducing computational costs compared to full-image and sliding window approaches.
Image forgery localization appears simplified case of semantic segmentation, facing perturbation of semantic objective content. 
Existing methods \cite{li2019localization,chen2021image,kwon2022learning,liu2022pscc} study traditional tampering techniques, and cannot adapt to the latest face manipulation algorithms. 
In this paper, we mainly focus on localizing regions manipulated by advanced face forgery techniques for real-fake mixed datasets.

\subsection{Noise and Frequency Forgery Clues}
To learn semantic-agnostic features, many image forgery localization approaches \cite{chen2021image,kwon2022learning,guo2023hierarchical} exploit noise or frequency artifacts to inhibit image content.
MVSS \cite{chen2021image} adopts BayarConv \cite{bayar2018constrained} to extract noise-view patterns on input RGB image. 
CAT-Net \cite{kwon2022learning} focuses on JPEG compression artifacts using the DCT coefficients segmentation model. 
HiFi-Net \cite{guo2023hierarchical} extracts RGB and frequency (LoG) features, but deep features may retain semantic information. 
In the face manipulation detection community, most methods \cite{durall2019unmasking,qian2020thinking,liu2021spatial,gu2021exploiting,luo2021generalizing,chen2021local} extract frequency/noise artifacts from RGB input. 
Other studies \cite{masi2020two,miao2022hierarchical,miao2023f} learn frequency forgery traces for detection but not localization tasks. 
We propose a multi-spectral class center module to suppress semantic content features in the deeper localization branch.

%-------------------------------------------------------------------------
\section{Methodology}

\subsection{Problem Formulation} \label{sec:method_1}
As demonstrated in Figure \ref{fig:framework} (a), our proposed face manipulation forensics architecture consists of a backbone network, a classification branch, and a localization branch, where the backbone network is utilized to project each input image $\mathcal{I} \in \mathbb{R}^{3 \times H \times W}$ into multi-scale feature space $\mathcal{F} = \{\mathcal{F}_{1}, \mathcal{F}_{2}, \mathcal{F}_{3}\}$, where $H \times W$ is the shape of the input image.
After that, a multi-level forgery patterns aggregation scheme is designed to aggregate $\mathcal{F}$ and output $\mathcal{F}_{A} \in \mathbb{R}^{C \times h \times w}$, where $C$ denotes for the number of feature channels.

Next, to exploit the global contextual representation of tampered regions over different frequency bands from aggregated $\mathcal{F}_{A}$, we propose the multi-spectral class center (MSCC) module as $\mathscr{M}$, and then we have: 
\begin{equation} \label{eq3}
   \mathcal{F}_{M} = \mathscr{M}(\mathcal{F}_{A}),
\end{equation}
where $\mathcal{F}_{M}  \in \mathbb{R}^{C \times h \times w}$ is the enhanced features from the perspective of centers of different spectral classes.
Finally, $\mathcal{F}_{M}$ is leveraged to predict the label of each pixel in the input image:
\begin{equation} \label{eq4}
   \mathcal{P}_1 = Upsample_{8\times}(\mathscr{C}_1 (\mathcal{F}_{M})),
\end{equation}
where $\mathscr{C}_1$ is a pixel-level classification head and $\mathcal{P}_1 \in \mathbb{R}^{k \times H \times W}$ indicates the predicted pixel-level class probability distribution.
Moreover, we apply the last layer output features $\mathcal{F}_{3}$ 
of the backbone network as image-level classification head $\mathscr{C}_2$ input, we have:
\begin{equation} \label{eq5}
   \mathcal{P}_2 = \mathscr{C}_2 (\mathcal{F}_{3}),
\end{equation}
in which, $\mathcal{P}_2 \in \mathbb{R}^{k}$ represents the image-level prediction probability distribution.
Here, $k$ is the number of classes and $k = 2$.

\subsection{Multi-level Features Aggregation}
\label{sec:method_2}
The forgery artifacts (\textit{e.g.}, blending boundary, checkboard, blur artifacts, etc.) and local structure are low-level texture features, which are mostly exiting shallow layers of the network \cite{liu2021spatial,luo2021generalizing}.
However, previous face manipulation localization methods \cite{nguyen2019multi,dang2020detection} primarily focused on deep semantic information and disregarded low-level texture features and location information, which would result in coarse and inaccurate output and disrupt some crucial low-level details (see Figure \ref{fig:vis_c40}).
To leverage the forgery-related low-level texture features, we propose the Multi-level Features Aggregation (MFA) scheme, which exploits texture-related information from multi-level and enhances the texture details of high-level semantic features.
 
As shown in Figure \ref{fig:framework} (b), we first gain multi-level features $\mathcal{F}_{1}, \mathcal{F}_{2}, \mathcal{F}_{3}$ from the backbone network and then employ three different aligned layers (\textit{i.e.}, $\mathscr{N}_1$, $\mathscr{N}_2$, and $\mathscr{N}_3$) for each of them:
\begin{equation} \label{eq6}
   \mathcal{F}^{'}_{1} = \mathscr{N}_1 (\mathcal{F}_{1}),
   \mathcal{F}^{'}_{2} = \mathscr{N}_2 (\mathcal{F}_{2}),
   \mathcal{F}^{'}_{3} = \mathscr{N}_3 (\mathcal{F}_{3}),
\end{equation}
where $\mathcal{F}^{'}_{1}, \mathcal{F}^{'}_{2}, \mathcal{F}^{'}_{3}  \in \mathbb{R}^{C \times h \times w}$.
Each aligned layer consists of a \textit{Conv} and a \textit{Downsample}, which aligns the different level features to assure the effectiveness of the lower-level texture information.
Then, we aggregate the aligned multi-level features $\mathcal{F}^{'}_{1}, \mathcal{F}^{'}_{2}, \mathcal{F}^{'}_{3}$ by channel-wise concatenation operation $Cat$ as follows:
\begin{equation} \label{eq7}
   \mathcal{F}_{A} = Conv(Cat ([\mathcal{F}^{'}_{1}, \mathcal{F}^{'}_{2}, \mathcal{F}^{'}_{3}])).
\end{equation}
where $\mathcal{F}_{A}  \in \mathbb{R}^{C \times h \times w}$ and the \textit{Conv} layer to make the channel size of $3C$ to $C$. 

\subsection{Multi-spectral Class Center}
\label{sec:method_3}
Previous face manipulation localization approaches \cite{nguyen2019multi,dang2020detection, jia2021inconsistency, nguyen2019multi, jia2021inconsistency, Huang2020FakeLocatorRL, Songsriin2019ComplementFF, Wu2022ExploringSF} have primarily focused on RGB features learning within the backbone network while neglecting to fully exploit the effectiveness of the rich forgery cues at the localization branch.
Other image manipulation localization approaches \cite{Wu2019ManTraNetMT, Zhou2018GenerateSA, Bappy2017ExploitingSS, Hao2021TransForensicsIF, chen2021image} apply noise or frequency information in the shallow layer of the backbone network, thus still keeping natural semantic object information in the localization branch.
As face manipulation localization models solely require the localization of tampered regions rather than all meaningful object regions, further analysis indicates that semantic objective features interfere with the forgery cues \cite{Hao2021TransForensicsIF,chen2021image}.
Therefore, the primary concern is how to develop and train a face manipulation localization model that can leverage frequency-aware forgery features with sensitivity towards manipulations in a deeper localization network.
The manipulated elements have discrepancies in the frequency domain compared to the authentic part, and extracting multi-frequency band information in the contextual features helps to suppress the semantic objective features \cite{qian2020thinking,liu2021spatial,chen2021local,luo2021generalizing}.
Inspired by these motivations, we propose a novel Multi-spectral Class Center (MSCC) module to learn semantic-agnostic forgery features from the different-frequency bands perspective, as shown in Figure \ref{fig:framework} (c).

\subsubsection{Discrete Cosine Transform Filters}
Following \cite{ahmed1974discrete,qin2021fcanet}, the 2D Discrete Cosine Transform (DCT) filters basis functions as follows:
\begin{equation} \label{eq8}
    \begin{aligned}
     D_{u,v} = \sum_{i=0}^{H-1} \sum_{j=0}^{W-1} d_{i,j} \cos(\dfrac{\pi u}{U}(i+\dfrac{1}{2}))\cos(\dfrac{\pi v}{V}(j+\dfrac{1}{2})) \\
     s.t. \;\; u \in \{0,1, \cdots,U-1\}, v \in \{0,1,\cdots,V-1\},
    \end{aligned}
\end{equation}
where $d \in \mathbb{R}^{H \times W}$ is a two-dimensional data and $D_{u,v} \in \mathbb{R}^{H \times W}$ is the 2D DCT frequency spectrum with the transformation basis of $(u,v)$. 
For simplicity, we define the above DCT operation as $\mathscr{D}_{n}(\cdot)$, in which $n \in \{0,1,\cdots, N-1\}$ and $N$ is the number of frequency transformation basis of $(u,v)$.
In this paper, we first split the features $\mathcal{F}_{A} \in \mathbb{R}^{C \times h \times w}$ into $N$ parts
along the channel dimension, where each channel of the $n$-th part feature $\mathcal{F}^{n}_{A} \in \mathbb{R}^{c \times h \times w}$ is defined $f^{n}_{i} \in \mathbb{R}^{h \times w}$, $i \in \{0,1,\cdots, c-1\}$ and $c = \frac{C}{N}$. 
Then, every $f^{n}_{i}$ is transformed through $\mathscr{D}_{n}(\cdot)$ with $n$-th transformation basis $(u,v)$, as follows:
\begin{equation} \label{eq9}
   \mathcal{\widetilde{F}}^{n}_{A} = Cat ([\mathscr{D}_{n}(f^{n}_{0}), \mathscr{D}_{n}(f^{n}_{1}), \cdots, \mathscr{D}_{n}(f^{n}_{c-1})]),
\end{equation}
where $\mathcal{\widetilde{F}}^{n}_{A} \in \mathbb{R}^{c \times h \times w}$ is the frequency features for specific spectral component.
Similarly, we can obtain the frequency information of the $\mathcal{F}_{A}$ for all spectral components and concatenate them together channel-wise:
\begin{equation} \label{eq10}
   \mathcal{\widetilde{F}}_{A} = Cat ([\mathcal{\widetilde{F}}^{0}_{A}, \mathcal{\widetilde{F}}^{1}_{A}, \cdots, \mathcal{\widetilde{F}}^{N-1}_{A}]),
\end{equation}
in which, $\mathcal{\widetilde{F}}_{A} \in \mathbb{R}^{(N \times c) \times h \times w}$ are frequency-aware feature maps with $N$ different frequency bands.
In the implementation, we set $N=64$ (i.e., $U=V=8$).

\subsubsection{Multi-spectral Class Center}
After getting the frequency feature maps $\mathcal{\widetilde{F}}_{A} \in \mathbb{R}^{ C \times h \times w}$, we split it along the channel dimension into $M$ different spectral features, denoted as $\mathcal{\widetilde{F}}_{S} \in \mathbb{R}^{M \times \frac{C}{M} \times h \times w}$.
Then we calculate the coarse segmentation predictions of different frequency components through a pixel-level classification head $\mathscr{C}_3$, then we have:
\begin{equation} \label{eq11}
   \mathcal{P}_{S} = \mathscr{C}_3 (\mathcal{\widetilde{F}}_{S}),
\end{equation}
where $\mathcal{P}_{S} \in \mathbb{R}^{M \times k \times h \times w}$ indicates the probability of a pixel belonging to a specific class in $M$ different spectral features, i.e., mapping the features from $\frac{C}{M}$ to $k$.
After that, we perform a matrix multiplication $\otimes$ between the $\mathcal{P}_{S}$ and the transpose of $\mathcal{\widetilde{F}}_{S}$ to calculate the multi-spectral class centers $\mathcal{F}_{class} \in \mathbb{R}^{M \times k \times \frac{C}{M}}$ as follows:
\begin{equation} \label{eq12}
   \mathcal{F}_{class} = \mathcal{P}_{S} \otimes \mathcal{\widetilde{F}}_{S}^{\top}.
\end{equation}
Multi-spectral class centers are expected to learn a global representation of each class from a different frequency perspective.
Since the class centers of the different spectra are calculated independently, there are missing interactions between them.
To address this, we first treat the multi-spectral class centers as distinct nodes, then message across each node, and finally update the features for each node.
The graph node modeling process can be formulated as follows:
\begin{equation} \label{eq13}
   \mathcal{F}^{'}_{class} = \mathscr{G} (\mathcal{F}_{class}),
\end{equation}
where $\mathscr{G}$ is a GCN layer that enhances the relationships between different spectral class centers.

\subsubsection{Feature Refinement}
We employ the multi-spectral class centers $\mathcal{F}^{'}_{class} \in \mathbb{R}^{M \times k \times \frac{C}{M}}$ to refine the aggregated multi-level features $\mathcal{F}_{A} $ through an attentional calculation mechanism.
We first compute a multi-spectral weight matrix to represent pixel similarity maps between each class center and the corresponding partial feature in $\mathcal{F}_{A}$, as follows:
\begin{equation} \label{eq14}
   \mathcal{W} = Softmax (\mathcal{F}_{A} \otimes (\mathcal{F}^{'}_{class})^{\top}),
\end{equation}
where $\mathcal{W} \in \mathbb{R}^{M \times hw \times k}$ and $\mathcal{F}_{A}$ is split by channel-wise and reshaped as $M \times hw \times \frac{C}{M}$.
Then, the weighted features $\mathcal{F}_{A}^{'} \in \mathbb{R}^{M \times hw \times \frac{C}{M}}$ are calculated as follows:
\begin{equation} \label{eq15}
   \mathcal{F}_{A}^{'} = \mathcal{W} \otimes \mathcal{F}^{'}_{class}.
\end{equation}
Finally, the multi-spectral class centers refined features $\mathcal{F}_{M}  \in \mathbb{R}^{C \times h \times w}$ is obtained by fusing the original features $\mathcal{F}_{A}$ and weighted features $\mathcal{F}_{A}^{'}$ via a \textit{Conv} layer, we have:
\begin{equation} \label{eq16}
   \mathcal{F}_{M} = Conv (Cat ([\mathcal{F}_{A}, \mathcal{F}_{A}^{'}])).
\end{equation}
Note that $\mathcal{F}_{A}^{'}$ is recovered and permuted to have a size of $C \times h \times w$ and the \textit{Conv} layer to make the channel size of $2C$ to $C$.

Our MSCC module represents pixel-class relationships over different spectra features.
The decomposed class centers are employed to calculate the attention of different frequency bands for suppressing semantic contextual information.
This is because the original semantic-aware features are frequency aliasing states,
with particularly low-frequency information dominating and high-frequency forgery cues easily discounted \cite{zhang2019making}, as shown in Fig. \ref{fig:motivation}.
Hence, our MSCC module enhances the capacity of the model to learn semantic-agnostic features that are sensitive to face manipulation traces.
In this way, the proposed MSCCNet effectively mitigates the disruption of deep semantic features in the localization decoder network, surpassing previous methods \cite{wang2021m2tr,chen2022self,nguyen2019multi,li2019localization,chen2021image,kwon2022learning,guo2023hierarchical}.

%--------------Figure dataset_vis---------------
\begin{figure*}[!t]
\centering
\includegraphics[width=0.95\linewidth]{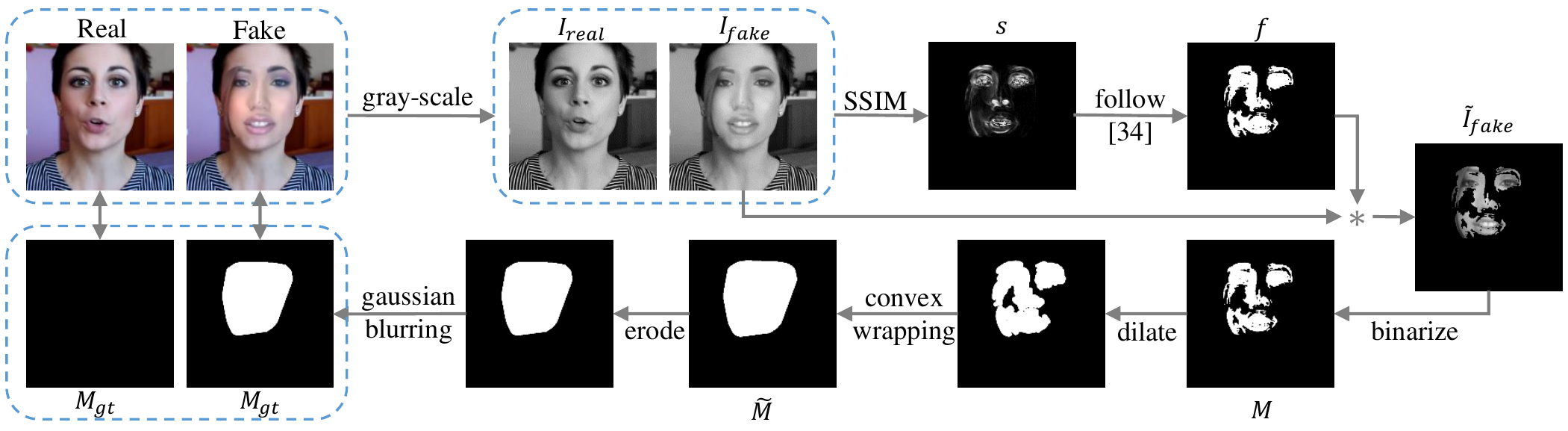}
\caption{Pixel-level annotation procedure for the P-FF++ dataset. The symbol $\ast$ is a multiplication operation.
}
\label{fig:dataset}
\end{figure*}
%--------------Figure dataset_vis---------------

\subsection{Objective Function}
\label{sec:method_4}
We first apply two cross-entropy loss functions for the predictions $\mathcal{P}_1$ and $\mathcal{P}_2$ of the MSCCNet, \textit{i.e.}, a pixel-level loss $\mathcal{L}_{seg}$ for localizing the manipulated regions and an image-level loss $\mathcal{L}_{cls}$ for classifying the authentic or manipulated face.
Then, for coarse segmentation predictions $\mathcal{P}_{S} \in \mathbb{R}^{M \times k \times h \times w}$ in Eq.(\ref{eq11}), we employ a \textit{$1 \times 1$ Conv} to fuse the multi-spectral results as follow:
\begin{equation} \label{eq17}
   \mathcal{P}_{S}^{'} = Conv (\mathcal{P}_{S}),
\end{equation}
where $\mathcal{P}_{S}^{'} \in \mathbb{R}^{k \times h \times w}$ is global representations. Similarly, the cross-entropy loss function is employed to calculate its loss $\mathcal{L}_{mscc}$.
Finally, the multi-task loss function $\mathcal{L}$ is used to jointly optimize the model parameters, we have:
\begin{equation} \label{eq18}
   \mathcal{L} = \mathcal{L}_{cls} + \mathcal{L}_{seg} + \mathcal{L}_{mscc}.
\end{equation}

%-------------------------------------------------------------------------
\section{Benchmark}
\label{sec:benchmark}

\subsection{Datasets} 
To facilitate the study of face manipulation localization, we define this task as recognizing pixel-level manipulated regions from a given face image.
We have selected the most widely-used P-FF++ dataset, and the Dolos \cite{Tantaru2024WACV} dataset which includes the latest diffusion-based generative models, to construct a comprehensive benchmark.

% %--------------Table Datasets---------------
% \begin{table}
% \caption{Details of the statistical quantity of the pixel-level FF++  \cite{rossler2019faceforensics++} datasets. 
% It includes both fake and corresponding real face images for each type of manipulation.
% The symbol $\star$ denotes the removal of duplicate real face images during the validation and testing phases.}
% \label{tab:dataset}
% \resizebox{0.99\linewidth}{!}{
% \begin{tabular}{ccccc}
% \toprule
% Types         & Train Set & Valid Set & Test Set & All Sets \\ \hline
% DF \cite{deepfake2019}    & $28,756$ & $5,560$ & $5,560$ & $39,876$    \\ 
% FF \cite{thies2016face2face}    & $28,724$ & $5,522$ & $5,560$ & $39,806$    \\ 
% FS \cite{faceswap2019}    & $28,760$ & $5,560$ & $5,560$ & $39,880$    \\ 
% FSh \cite{li2019faceshifter}  & $28,760$ & $5,560$ & $5,560$ & $39,880$    \\ 
% NT \cite{thies2019deferred}  & $28,724$ & $5,522$ & $5,560$ & $39,806$    \\ 
% All Types     & $143,724$ & ${16,922}^{\star}$ & ${16,929}^{\star}$ & $177,575$    \\ 
% \bottomrule
% \end{tabular}
% }
% \end{table}      
% %--------------Table Datasets---------------

%--------------Table Datasets---------------
\begin{table}
\begin{center}
\caption{Details of the statistical quantity of the P-FF++  \cite{rossler2019faceforensics++} and Dolos datasets. 
The P-FF++ includes both fake and corresponding real face images for each type of manipulation.
The symbol $\star$ denotes removing duplicate real face images during the validation and testing phases.}
\label{tab:dataset}
\resizebox{0.99\linewidth}{!}{
\begin{tabular}{cccccc}
\toprule
\multicolumn{2}{c}{Types} & Train Set & Valid Set & Test Set & All Sets \\ \hline
\multirow{6}{*}{P-FF++}
& DF \cite{deepfake2019} + Real    & $28,756$ & $5,560$ & $5,560$ & $39,876$    \\ 
& FF \cite{thies2016face2face} + Real    & $28,724$ & $5,522$ & $5,560$ & $39,806$    \\ 
& FS \cite{faceswap2019} + Real    & $28,760$ & $5,560$ & $5,560$ & $39,880$    \\ 
& FSh \cite{li2019faceshifter} + Real  & $28,760$ & $5,560$ & $5,560$ & $39,880$    \\ 
& NT \cite{thies2019deferred} + Real  & $28,724$ & $5,522$ & $5,560$ & $39,806$    \\ 
& All Types     & $143,724$ & ${16,922}^{\star}$ & ${16,929}^{\star}$ & $177,575$    \\ 
\hline
\multirow{6}{*}{Dolos}
& Real \cite{karras2017ProgressiveGO}    & $9,000$ & $900$ & $900$ & $10,800$ \\
& Repaint \cite{lugmayr2022RePaint}    & $30,000$ & $3,000$ & $8,500$ & $41,500$    \\ 
& LDM \cite{suvorov2021resolution}    & $9,000$ & $900$ & $900$ & $10,800$    \\ 
& LaMa \cite{rombach2022high}  & $9,000$ & $900$ & $900$ & $10,800$    \\ 
& Pluralistic \cite{zheng2019pluralistic}  & $9,000$ & $900$ & $900$ & $10,800$    \\ 
& All Types     & $76,000$ & $7,600$ & $12,100$ & $95,700$    \\ 

\bottomrule
\end{tabular}
}
\end{center}
\end{table}      
%--------------Table Datasets---------------
% Repaint \cite{lugmayr2022RePaint}, LaMa \cite{rombach2022high}, LDM \cite{suvorov2021resolution}, and Pluralistic \cite{zheng2019pluralistic}

\subsubsection{Pixel-level Annotation for FaceForensics++ (P-FF++)} \label{sec:dataset_2}
The FF++ \cite{rossler2019faceforensics++} is a challenging face forgery video dataset with 1,000 original YouTube videos and 5,000 corresponding fake videos generated through 5 manipulation methods: Deepfakes (DF) \cite{deepfake2019}, Face2Face (FF) \cite{thies2016face2face}, FaceSwap (FS) \cite{faceswap2019}, FaceShifter (FSh) \cite{li2019faceshifter}, and NeuralTextures (NT) \cite{thies2019deferred}. It includes 3 quality levels: Raw (C0), High (C23), and Low (C40).

We further preprocess FF++ \cite{rossler2019faceforensics++} with annotations for forged region localization tasks. We apply real-fake image pairs from FF++ to generate pixel-level annotation, as forgery images and corresponding authentic images have pixel-level differences in manipulated regions and are identical elsewhere \cite{dang2020detection,chen2021local,wang2021m2tr, Zhao2021learning,li2020face}. 
As shown in Fig. \ref{fig:dataset}, for real and fake RGB image pairs, we convert them to grayscale, and compute SSIM \cite{Wang2004ImageQA} between them to produce SSIM map $S$ in the range of $[0,1]$, following \cite{Zhao2021learning}. 
To accurately portray pixel-level discrepancy $S$ on forged images, we first compute coarse manipulated regions factor $f$ from $S$. 
Second, we multiply $f$ and fake image to obtain binarized $M$. As $M$ is scattered, we dilate it, apply convex wrapping twice for comprehensive tamper region mask, then erode edges and apply Gaussian blurring to generate ground truth masks  $M_{gt}$ for fake images. For corresponding real images, we apply zero-maps as  $M_{gt}$.

For each P-FF++ video, we select up to 20 frames to form single-face manipulation image datasets and obtain forged region labels using the proposed annotation procedure. We divide training, validation, and testing sets following the original work. Detailed statistics of pixel-level FF++ (P-FF++) are shown in Table \ref{tab:dataset}.

\subsubsection{Dolos}
Dolos \cite{Tantaru2024WACV} is a recently released dataset for local face manipulation. The real face images are sourced from CelebA-HQ \cite{karras2017ProgressiveGO}, and four different forgery techniques were applied: diffusion-based Repaint \cite{lugmayr2022RePaint} and LDM \cite{rombach2022high}, Fourier convolution-based LaMa  \cite{suvorov2021resolution}, and GAN-based Pluralistic \cite{zheng2019pluralistic}. 
It should be noted that this dataset retains the mask annotations of the forgery regions during the creation process.
The dataset details are summarized in Table \ref{tab:dataset}.

\subsection{Baseline Methods}
We conduct a competitive benchmark for face manipulation localization, including quantitative and qualitative evaluations, across various scenarios. For fair and reproducible comparison, we broadly select publicly available methods associated with localizing tampered faces:
(1) Face manipulation detection with segmentation loss \cite{wang2021m2tr,chen2022self}.  
(2) Face manipulation localization methods \cite{nguyen2019multi,dang2020detection}.
(3) Image forgery localization methods \cite{li2019localization,chen2021image,kwon2022learning,guo2023hierarchical}.

Baseline methods described:
(1) HPFCN \cite{li2019localization}: High-pass filtered fully convolutional network for deep inpainting localization, lacks classification branch.
(2) Multi-task \cite{nguyen2019multi}: Multi-task network with encoder, Y-shaped decoder for classification, segmentation, and reconstruction.
(3) FFD \cite{dang2020detection}: Utilizes attention to process feature maps for classification and localization highlights.
(4) M2TR \cite{wang2021m2tr}: Multi-scale patches for local inconsistencies, cross-modality for multi-modal fusion. Applies segmentation loss but lacks pixel-level results.
(5) SLADD \cite{chen2022self}: Large forgery augmentation space, adversarial training, forgery region prediction head for classification.
(6) MVSS \cite{chen2021image}: Multi-view feature learning, multi-scale supervision for manipulation localization.
(7) CATNet \cite{kwon2022learning}: Focuses on JPEG artifacts, not suitable for FF++ video compression.
(8) HiFi-Net \cite{guo2023hierarchical}: Color, frequency blocks for generation artifacts, multi-branch feature extractor.

M2TR \cite{wang2021m2tr} and SLADD \cite{chen2022self} are designed for Deepfake detection, but we compute their pixel-level results. HPFCN \cite{li2019localization} and MVSS \cite{chen2021image} utilize a backbone for additional classification branches like our MSCCNet.

\subsection{Metrics}
The Accuracy (ACC) and Area Under the Receiver Operating Characteristic Curve (AUC) are reported for face manipulation detection comparison metrics. 
For the evaluation of localization results, we employ the pixel-level F1-score and mIoU (mean of class-wise intersection over union).
Unless otherwise noted, the computed evaluation metrics are reported for both authentic and manipulated images.
The higher value indicates that the performance is better.

\subsection{Evaluation Protocols} \label{sec:experiment_2}
To completely evaluate our MSCCNet, we adopt three evaluation protocols: 
1) Intra-dataset: We adopt the C23 and C40 of the P-FF++, and Dolos \cite{Tantaru2024WACV} for intra-test evaluation. 
2) Unseen datasets (cross-datasets): We train the proposed method on the C23 set of the P-FF++ dataset and then test it on the unseen test set of Dolos \cite{Tantaru2024WACV}, and vice versa.
3) Unseen manipulations (cross-manipulation): We perform experiments on the C40 set of the P-FF++ dataset through a leave-one-out strategy. 
Specifically, there are five manipulation types of fake face images in the C40 set, one type is used as a test set while the remaining four types form the training set. 
For the Dolos \cite{Tantaru2024WACV} dataset, we train on the Repaint \cite{lugmayr2022RePaint} set and then test the other three manipulation sets (i.e., LaMa \cite{suvorov2021resolution}, LDM \cite{rombach2022high}, and Pluralistic \cite{zheng2019pluralistic}).

%-------------------------------------------------------------------------
\section{Experiments}

\subsection{Experimental Setup} \label{sec:experiment_1}
Our MSCCNet's backbone is dilated ResNet-50 \cite{He2015DeepRL}, the classification branch is a simple MLP layer, and the localization branch consists of proposed MFA and MSCC modules. 
Specifically, the ResNet-50 backbone is initialized with ImageNet pre-trained weights, while the remaining layers/modules are randomly initialized. 
Dilated ResNet-50's output stride is 8, so h=H/8 and w=W/8 in MSCCNet. The remaining benchmark models follow original papers unless otherwise stated.

We train MSCCNet with SGD: initial learning rate $0.009$, momentum$ 0.9$, weight decay $5e-4$. 
Learning rate decays by poly policy $(1 - \frac{iter}{total_iter})^{0.9}$. Input size 512x512, batch size 64. 
Random horizontal flipping for data augmentation. 
Synchronized batch norm by PyTorch 1.8.1 for multi-GPU training. 
Other methods follow original papers unless stated.

\begin{table*}
\begin{center}
\caption{Intra-dataset results for face manipulation localization and detection on the P-FF++ and Dolos datasets. The C40 and C23 indicate different compression levels.}
\label{tab:intra_test}
\resizebox{0.99\linewidth}{!}{
\begin{tabular}{clcccc|cccc|cccc}
\toprule
\multirow{3}{*}{Methods} &
  \multicolumn{1}{c}{\multirow{3}{*}{References}} &
  \multicolumn{4}{c|}{P-FF++ (C40)} &
  \multicolumn{4}{c|}{P-FF++ (C23)} &
  \multicolumn{4}{c}{Dolos} \\ 
  \cline{3-14} 
 &
  \multicolumn{1}{c}{} &
  \multicolumn{2}{c}{Image-level} &
  \multicolumn{2}{c|}{Pixel-level} &
  \multicolumn{2}{c}{Image-level} &
  \multicolumn{2}{c|}{Pixel-level} &
  \multicolumn{2}{c}{Image-level} &
  \multicolumn{2}{c}{Pixel-level} \\ \cline{3-14} 
           & \multicolumn{1}{c}{} & ACC   & AUC   & F1    & mIoU  & ACC   & AUC   & F1    & mIoU & ACC   & AUC   & F1    & mIoU  \\ \midrule
HPFCN \cite{li2019localization}   & ICCV 2019 & 82.05 & 69.08 & 60.53 & 48.37 & 86.60 & 88.69 & 69.12 & 55.91 & 82.46 & 90.18 & 87.36 & 78.62 \\
Muilt-task \cite{nguyen2019multi} & BTAS 2019 & 71.09 & 74.05 & 74.91 & 61.86 & 88.08 & 93.41 & 81.88 & 70.39 & 95.27 & 97.86 & 87.72 & 79.22 \\
FFD \cite{dang2020detection}      & CVPR 2020 & 81.65 & 80.05 & 61.24 & 48.84 & 90.94 & 94.54 & 72.63 & 59.27 & 89.05 & 78.05 & 45.42 & 41.61 \\
M2TR \cite{wang2021m2tr}          & ICMR 2022 & 86.18 & 86.34 & 75.33 & 62.02 & 93.44 & 97.18 & 85.13 & 74.78 & 98.48 & 99.86 & 89.29 & 81.51 \\
SLADD \cite{chen2022self}         & CVPR 2022 & 86.25 & 85.53 & 70.95 & 57.86 & 91.12 & 97.23 & 79.96 & 67.87 & 87.04 & 89.08 & 38.54 & 27.20 \\
MVSS \cite{chen2021image}         & TPAMI 2022 & 85.08 & 81.99 & 82.34 & 70.82 & 95.30 & 98.71 & 88.79 & 80.20 & 96.10 & 99.11 & 95.44 & 91.44 \\
CAT-Net \cite{kwon2022learning}   & IJCV 2022 & 85.86 & 85.77 & 84.89 & 74.40 & 96.14 & 98.83 & 89.18 & 80.86 & 96.74 & 99.84 & 96.07 & 92.56 \\
HiFi-Net \cite{guo2023hierarchical}  & CVPR 2023 & 72.77 & 80.28 & 76.66 & 63.17 & 89.46 & 97.35 & 84.81 & 74.28 & 98.43 & \textbf{99.98} & 93.19 & 87.56 \\ \midrule
\textbf{MSCCNet (ours)} & --      & \textbf{88.07} & \textbf{87.61} & \textbf{86.82} & \textbf{77.22} & \textbf{97.21} & \textbf{98.94} & \textbf{90.71} & \textbf{83.29} & \textbf{99.02} & 99.96 & \textbf{98.93} & \textbf{97.89} \\ 
\bottomrule
\end{tabular}
}
\end{center}
\end{table*}
%--------------Table Intra-datasets---------------

\subsection{Intra-dataset Evaluation}
We first investigate the localization performance of benchmark approaches on the P-FF++ and Dolos \cite{Tantaru2024WACV} datasets.
This task is more practical and challenging, yet is rarely explored in the previous literature.
As shown in Table \ref{tab:intra_test}, the FFD \cite{dang2020detection} model exhibits inadequate localization results due to its utilization of low-resolution attention maps as prediction masks. Furthermore, it lacks the ability to incorporate global contextual representation in its localization branch, rendering it unsuitable for forgery localization tasks.
M2TR \cite{wang2021m2tr} and SLADD \cite{chen2022self} apply the semantic segmentation pipeline to supervise the manipulated regions
but their main objective is to enhance detection performance rather than localization, consequently leading to unsatisfactory localization performance.
Another crucial factor is the disregard for the negative impact of semantic objective information in these methods \cite{nguyen2019multi,dang2020detection,wang2021m2tr,chen2022self}.
In the image forgery localization community, while some approaches \cite{li2019localization, chen2021image, kwon2022learning,guo2023hierarchical} address this drawback and achieve notable performance improvements from a noise or frequency perspective, they still struggle to effectively suppress semantic objective information in the deep features of the localization branch.
For example, HPFCN \cite{li2019localization} employs a filter on the input RGB image and only achieves a $60.53$ F1-score. 
MVSS \cite{chen2021image}, CAT-Net \cite{kwon2022learning}, and HiFi-Net \cite{guo2023hierarchical} fuse the features of the RGB image with noise- or frequency-view patterns, but they also extract them on the inputs.
Hence, their localization performance on face manipulation datasets is poorer than our MSCCNet, especially on the FF++ C40 dataset \cite{rossler2019faceforensics++}.
This is inherently caused by the diminished discrepancy between tampered and real areas in low-quality forged images, leading to a reduction in distinctive semantic objective features and consequent localization failures.
In comparison to alternative models, our MSCCNet model exhibits superior performance, especially on the C40 dataset. 
This outcome suggests that the proposed MFA and MSCC modules enhance global contextual representations that are multi-frequency forgery features while enabling the suppression of objective semantic-related information.

We next analyze the image-level classification performance of the face forgery localization approaches on the P-FF++ and Dolos \cite{Tantaru2024WACV} datasets. 
Face manipulation detection methodologies \cite{wang2021m2tr,chen2022self} have already extensively studied classification tasks, so they achieve remarkable results.
As shown in Table \ref{tab:intra_test}, these classification results of C40 sets show that FFD \cite{dang2020detection} and Multi-task \cite{nguyen2019multi} are not suitable for low-quality datasets.
Our findings from Table \ref{tab:intra_test} illustrate that preceding image forgery localization methods \cite{li2019localization, guo2023hierarchical} have yielded inadequate classification outcomes on the C40 and C23 datasets.
Our MSCCNet outperforms all benchmark methods in terms of ACC and AUC on the C40 set.
It is worth noting that the proposed MFA and MSCC modules are specifically designed to enhance the localization branch's function, without directly augmenting image-level classification abilities.

\begin{table}
\begin{center}
\caption{Generalization to unseen datasets. The model is trained on the training set of P-FF++ C23 while tested on the test set of Dolos, and vice versa.
} 
\label{tab:cross_dataset}
\resizebox{0.99\linewidth}{!}{
\begin{tabular}{ccccc|cc}
\toprule
\multirow{2}{*}{Methods} &
  \multicolumn{2}{c}{C23 $\rightarrow$ Dolos} &
  \multicolumn{2}{c|}{Dolos $\rightarrow$ C23} 
  & \multicolumn{2}{c}{\textit{Average}} \\ 
  \cline{2-7} 
                                    & F1    & mIoU  & F1    & mIoU  & \textit{F1}    & \textit{mIoU} \\ \midrule
HPFCN \cite{li2019localization}     & 45.53 & 41.62 & 43.82 & 36.90 & \textit{44.68} & \textit{39.26} \\
Muilt-task \cite{nguyen2019multi}   & 48.62 & 43.25 & 42.22 & 36.38 & \textit{45.42} & \textit{39.82} \\ 
FFD \cite{dang2020detection}        & 45.75 & 41.73 & 42.08 & 36.33 & \textit{43.92} & \textit{39.03} \\
M2TR \cite{wang2021m2tr}            & 59.32 & 48.40 & 42.73 & 36.89 & \textit{51.03} & \textit{42.65} \\
SLADD \cite{chen2022self}           & 49.58 & 43.70 & 37.38 & 23.01 & \textit{43.48} & \textit{33.36} \\
MVSS \cite{chen2021image}           & 59.57 & 46.89 & 56.76 & 43.21 & \textit{58.17} & \textit{45.05} \\
CAT-Net \cite{kwon2022learning}     & 56.30 & 46.10 & 55.48 & 43.51 & \textit{55.89} & \textit{44.81} \\
HiFi-Net \cite{guo2023hierarchical} & 47.10 & 42.37 & 56.08 & 39.05 & \textit{51.59} & \textit{40.71} \\ \midrule
\textbf{MSCCNet (ours)}             & 59.84 & 47.88 & 58.71 & 45.64 & \textbf{\textit{59.28}} & \textbf{\textit{46.76}} \\ 
\bottomrule
\end{tabular}
}
\end{center}
\end{table}
\begin{table*}
\begin{center}
\caption{Generalization to unseen manipulations on the P-FF++ C40 dataset, which consists of five manipulation methods. We train on four methods and test on the other one method. The italicized numbers indicate the average of the five different generalization results.
The F1 and mIoU are pixel-level results.
}
\label{tab:cross_4to1_c40}
\resizebox{0.99\linewidth}{!}{
\begin{tabular}{ccccccccccc|cc}
\toprule
\multirow{2}{*}{\begin{tabular}[c]{@{}c@{}}Methods\end{tabular}} & \multicolumn{2}{c}{Deepfakes} & \multicolumn{2}{c}{Face2Face} & \multicolumn{2}{c}{FaceSwap} & \multicolumn{2}{c}{FaceShifter} & \multicolumn{2}{c|}{NeuralTextures} & \multicolumn{2}{c}{\textit{Average}} \\ \cline{2-13} 
  & F1 & mIoU & F1 & mIoU & F1 & mIoU & F1 & mIoU & F1 & mIoU & \textit{F1} & \textit{mIoU} \\ \midrule
HPFCN \cite{li2019localization}    & 66.09 & 55.78 & 56.79 & 47.06 & 63.47 & 53.69 & 57.16 & 46.61 & 53.31 & 44.54 & \textit{59.53} & \textit{49.54} \\ 
Multi-task \cite{nguyen2019multi}  & 72.74 & 61.17 & 60.77 & 50.04 & 57.25 & 48.54 & 58.21 & 47.75 & 53.42 & 44.88 & \textit{60.48} & \textit{50.48} \\ 
FFD \cite{dang2020detection}       & 67.77 & 56.54 & 46.24 & 41.06 & 53.24 & 47.50 & 61.31 & 49.69 & 54.50 & 45.13 & \textit{56.61} & \textit{47.98} \\
M2TR \cite{wang2021m2tr}           & 73.15 & 60.83 & 62.73 & 51.17 & 64.21 & 52.89 & 54.02 & 44.91 & 57.58 & 47.11 & \textit{62.34} & \textit{51.38} \\
SLADD \cite{chen2022self}          & 74.79 & 63.58 & 59.91 & 49.31 & 63.08 & 53.71 & 58.82 & 48.11 & 55.80 & 46.12 & \textit{62.48} & \textit{52.17} \\
MVSS \cite{chen2021image}        & 70.61 & 58.11 & 62.07 & 50.71 & 63.52 & 53.03 & 62.55 & 49.75 & 55.36 & 45.61 & \textit{62.82} & \textit{51.44} \\
CAT-Net \cite{kwon2022learning}    & 71.13 & 58.32 & 63.92 & 51.51 & 64.55 & 53.07 & 63.76 & 51.40 & 56.89 & 47.19 & \textit{64.05} & \textit{52.30} \\
HiFi-Net \cite{guo2023hierarchical} & 52.41 & 44.94 & 61.35 & 50.15 & 60.74 & 49.83 & 52.64 & 44.12 & 55.26 & 45.97 & \textit{56.48} & \textit{47.00} \\ \midrule
\textbf{MSCCNet (ours)}            & 75.28 & 63.20 & 64.23 & 52.21 & 65.15 & 54.03 & 63.81 & 51.58 & 57.79 & 47.37 & \textit{\textbf{65.25}} & \textit{\textbf{53.68}} \\ 
\bottomrule
\end{tabular}
}
\end{center}
\end{table*}
%--------------Table Unseen Manipulation---------------

%--------------Table Unseen Manipulation---------------
\begin{table}
\begin{center}
\caption{Generalization to unseen manipulations on the Dolos dataset which consists of four manipulation methods. We train on the Repaint set and test on the other three sets.
The F1 and mIoU are pixel-level results.
}
\label{tab:cross_4to1_dolos}
\resizebox{0.99\linewidth}{!}{
\begin{tabular}{ccccccc|cc}
\toprule
\multirow{2}{*}{\begin{tabular}[c]{@{}c@{}}Methods\end{tabular}} & \multicolumn{2}{c}{LaMa} & \multicolumn{2}{c}{LDM} & \multicolumn{2}{c|}{Pluralistic} & \multicolumn{2}{c}{\textit{Average}} \\ \cline{2-9} 
  & F1 & mIoU & F1 & mIoU & F1 & mIoU & \textit{F1} & \textit{mIoU} \\ \midrule
HPFCN \cite{li2019localization}    & 56.99 & 48.40 & 53.54 & 43.07 & 49.42 & 40.72 & \textit{53.32} & \textit{44.06} \\ 
Multi-task \cite{nguyen2019multi}  & 61.65 & 47.91 & 52.33 & 37.98 & 74.82 & 62.79 & \textit{62.93} & \textit{49.56} \\ 
FFD \cite{dang2020detection}       & 48.02 & 45.31 & 58.93 & 51.37 & 48.44 & 45.46 & \textit{51.80} & \textit{47.38} \\
M2TR \cite{wang2021m2tr}           & 62.73 & 51.17 & 64.21 & 52.89 & 54.02 & 44.91 & \textit{60.32} & \textit{49.66} \\
SLADD \cite{chen2022self}          & 20.01 & 11.15 & 19.75 & 11.02 & 20.68 & 11.59 & \textit{20.15} & \textit{11.25} \\
MVSS \cite{chen2021image}          & 93.51 & 88.28 & 53.02 & 38.64 & 80.22 & 69.88 & \textit{75.58} & \textit{65.60} \\
CAT-Net \cite{kwon2022learning}    & 80.14 & 69.29 & 52.22 & 37.87 & 93.30 & 87.94 & \textit{75.22} & \textit{65.03} \\
HiFi-Net \cite{guo2023hierarchical} & 55.81 & 44.31 & 53.34 & 40.97 & 65.11 & 53.21 & \textit{58.08} & \textit{46.16} \\ \midrule
\textbf{MSCCNet (ours)}            & 98.47 & 97.01 & 55.74 & 41.88 & 91.54 & 85.16 & \textbf{\textit{81.92}} & \textbf{\textit{74.68}} \\ 
\bottomrule
\end{tabular}
}
\end{center}
\end{table}
%--------------Table Unseen Manipulation---------------

\subsection{Unseen Datasets Evaluation}
The unseen datasets are created by anonymous forgery methodology based on unknown source data. 
As shown in Table \ref{tab:cross_dataset}, we conduct cross-dataset experiments to evaluate the generalization capacity of the face manipulation localization models on unseen C23 or Dolos \cite{Tantaru2024WACV} datasets.  
Regarding the localization results for face manipulation, CAT-Net \cite{kwon2022learning} and MVSS \cite{chen2021image} accomplish significant performance by learning semantic-agnostic features. However, their localization branch networks do not fully excel in this aspect.
Exiting face forgery detection and localization methods have not fully taken into account the frequency-related forgery features of the localization branch network. 
In contrast, the proposed MSCCNet effectively inhibits the image semantic content of deeper features by utilizing multi-spectral class centers, thereby achieving this target.
From Table \ref{tab:cross_dataset}, our MSCCNet significantly outperforms all the competitors, which suggests that multi-spectral forgery cues offer a significant contribution to generalization.

\subsection{Unseen Manipulation Evaluation}
To assess the cross-manipulation generalization capabilities of different face manipulation localization models, we conduct the unseen manipulation evaluation experiments in Table \ref{tab:cross_4to1_c40} and \ref{tab:cross_4to1_dolos}.
These results demonstrate that our MSCCNet achieves exceptional localization generalization performance ($65.25\%$ F1-score and $53.68\%$ mIoU) to novel forgeries, surpassing most approaches. 
Despite the various manipulation methods employed in the five types of manipulations (Deepfakes \cite{deepfake2019}, Face2Face \cite{thies2016face2face}, FaceSwap \cite{faceswap2019}, FaceShifter \cite{li2019faceshifter}, and NeuralTextures \cite{thies2019deferred}) within the P-FF++ dataset, each of which focuses on different tasks, the proposed MSCCNet succeeded in learning a generalized discriminative feature on four of the manipulations and generalized to the remaining one. 
Different types of forgeries exhibit varying levels of difficulty, with Deepfakes \cite{deepfake2019} generally being easier to localize compared to NeuralTextures \cite{thies2019deferred} forgeries, which are often more challenging.
As shown in Table \ref{tab:cross_4to1_dolos}, 
our MSCCNet also outperforms other methods in terms of average localization performance, indicating that our model can adapt to diffusion-based synthetic face data.

%--------------Figure Visualization---------------
\begin{figure*}[!t]
\centering
\includegraphics[width=0.99\textwidth]{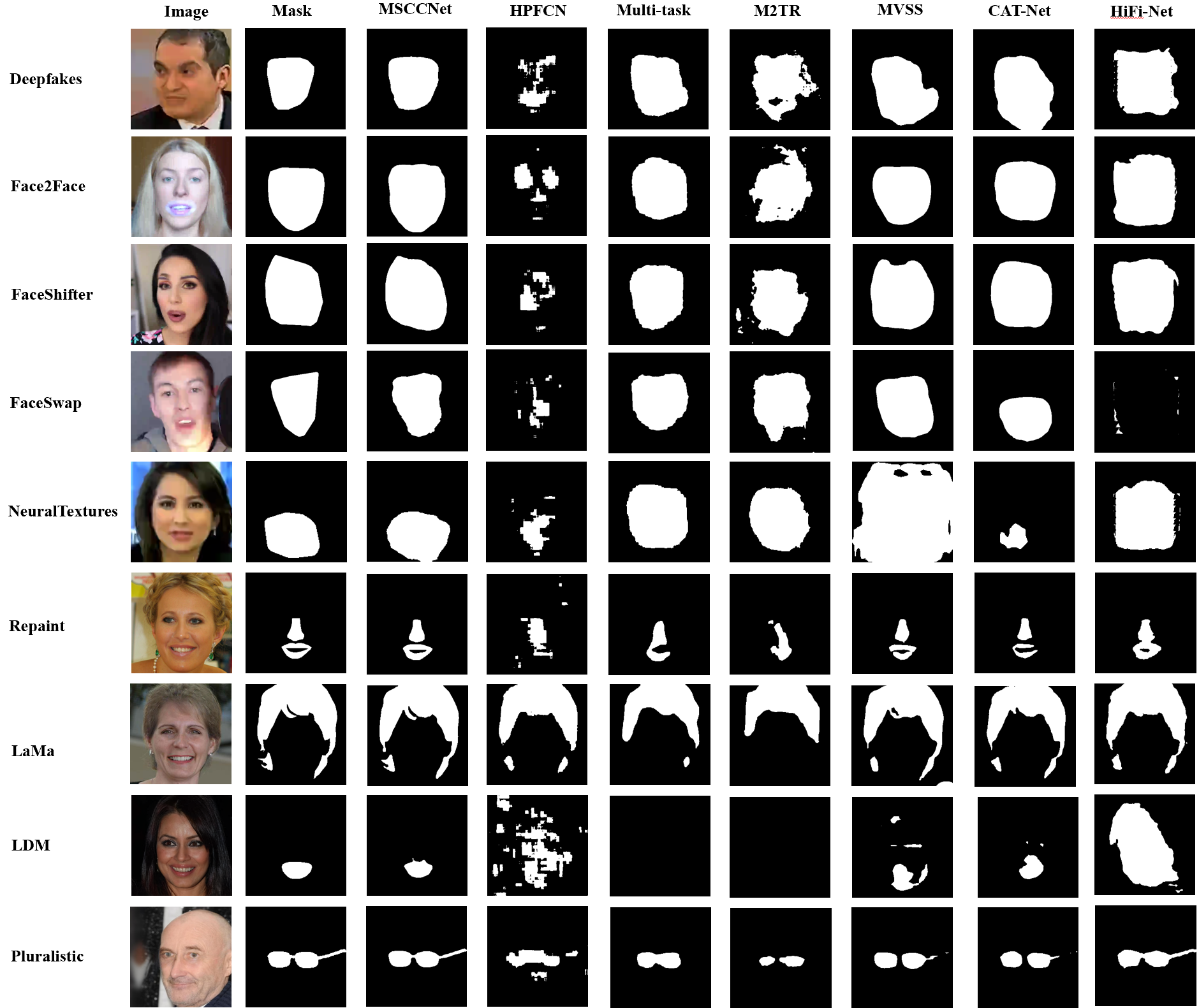}
\caption{Visualization mask predictions of baseline methods and our MSCCNet. The examples are randomly selected from the C40 test set of P-FF++ and Dolos. Every row indicates different face manipulation technologies. 
Columns Image and Mask represent the input forged face and its corresponding pixel-level label, respectively.
}
\label{fig:vis_c40}
\end{figure*}
%--------------Figure Visualization---------------

\subsection{Qualitative Comparisons}
After training, our model can generate high-quality mask predictions that depict tampering locations on the test set.
Here, we provide some qualitative samples in Figure \ref{fig:vis_c40}.
Since the FFD \cite{dang2020detection} and SLADD \cite{chen2022self} rely on low-resolution attention maps, their predictions tend to be small and coarse-grained. 
Therefore, we did not display and compare their predicted masks.
HPFCN \cite{li2019localization} proves inadequate for advanced face manipulation images, thus failing to predict tampered regions accurately.
The detrimental effects of Multi-task \cite{nguyen2019multi} and M2TR \cite{wang2021m2tr}, which excessively prioritize objective semantic features, are evident in Figure \ref{fig:vis_c40}.
For example, NeuralTextures \cite{thies2019deferred} is local forgery technology, while Multi-task \cite{nguyen2019multi} and M2TR \cite{wang2021m2tr} predict the whole face object regions.
In the case of LDM \cite{rombach2022high} row, where the manipulated region is only mouth, MVSS \cite{chen2021image}, CAT-Net \cite{kwon2022learning}, and HiFi-Net \cite{guo2023hierarchical} exhibit localization errors.
Meanwhile, Multi-task \cite{nguyen2019multi} and M2TR \cite{wang2021m2tr} are unable to predict the tampered regions, and instead consider them to be part of the original image.
The superior performance of our MSCCNet is evident from its ability to identify tampered regions in distinct types of forgeries.
This capability highlights the strength of our method in effectively modeling frequency-related forgery features in the localization network.

% %--------------Table Extend Experiment---------------
% \begin{table}
% \begin{center}
% \caption{Extend experiment on image forgery datasets. The model is trained on the CASIAv2 \cite{casiav2} dataset while tested on the CASIAv1 \cite{casiav1} and COVER \cite{2016COVERAGE} datasets. The default decision threshold of $0.5$ is used for all models, following MVSS \cite{chen2021image}. 
% }
% \label{tab:image_forgery}
% \resizebox{0.99\linewidth}{!}{
% \begin{tabular}{ccc|cc}
% \toprule
% \multirow{2}{*}{Methods} & \multicolumn{2}{c|}{Pixel-level F1} & \multicolumn{2}{c}{Image-level AUC} \\ \cline{2-5} 
%            & CASIAv1 & COVER  & CASIAv1 & COVER  \\ \midrule
% ManTra-Net \cite{Wu2019ManTraNetMT} & 15.5   & 28.6    & 14.1   & 54.3  \\
% CR-CNN \cite{yang2020constrained}   & 40.5   & 29.1    & 76.6   & 54.6  \\
% GSR-Net \cite{zhou2020generate}     & 38.7   & 28.5    & 50.2   & 45.6  \\
% MVSS \cite{chen2021image}          & 45.2   & 45.3     & 83.9   & 57.3  \\ \midrule
% \textbf{MSCCNet(ours)}             & \textbf{49.6} & \textbf{49.2} & \textbf{85.8}  & \textbf{58.6} \\ 
% \bottomrule
% \end{tabular}
% }
% \end{center}
% \end{table}
% %--------------Table Extend Experiment---------------

%--------------Table Extend Experiment---------------
\begin{table}
\begin{center}
\caption{Extend experiment on image forgery datasets. The model is trained on the CASIAv2 \cite{casiav2} dataset while tested on the other five datasets. The evaluation metric is the pixel-level F1 score.
}
\label{tab:image_forgery}
\resizebox{0.99\linewidth}{!}{
\begin{tabular}{ccccccc|c}
\toprule
Methods & References & COVER & Columbia & NIST16 & CASIAv1 & IMD2020 & \textit{Average} \\ \midrule
Mantra-Net \cite{Wu2019ManTraNetMT} & CVPR 2019 & 09.0 & 24.3 & 10.4 & 12.5 & 05.5 & \textit{12.3} \\
MVSS-Net \cite{chen2021image} & TPAMI 2022 & 25.9 & 38.6 & 24.6 & 53.4 & 27.9 & \textit{34.1} \\
ObjectFormer \cite{wang2022objectformer}   & CVPR 2022 & 29.4 & 33.6 & 17.3 & 42.9 & 17.3 & \textit{28.1} \\
PSCC-Net \cite{liu2022pscc} & TCSVT 2022 & 23.1 & 60.4 & 21.4 & 37.8 & 23.5 & \textit{33.3} \\
NCL-IML \cite{zhou2023pre} & ICCV 2023 & 22.5 & 44.6 & 26.0 & 50.2 & 23.7 & \textit{33.4} \\ \midrule
\textbf{MSCCNet(ours)} & -- & 24.14 & 37.04 & 22.52 & 65.37 & 27.63 & \textbf{\textit{35.34}} \\
\bottomrule
\end{tabular}
}
\end{center}
\end{table}
%--------------Table Extend Experiment---------------

\subsection{Extend Experiment} \label{sec:experiment_2_1}
To further validate the effectiveness of our approach, we conducted experiments on the image forgery datasets following \cite{ma2024imdlbenco}.
In general, the models are trained on the CASIAv2 \cite{casiav2} dataset, and tested on five unseen testing datasets including CASIAv1 \cite{casiav1}, COVER \cite{2016COVERAGE}, IMD2020 \cite{novozamsky2020imd2020}, NIST16 \cite{guan2019mfc}, and Columbia \cite{hsu2006detecting}.

As presented in Table \ref{tab:image_forgery}, the localization results for other methods are obtained from the \cite{ma2024imdlbenco} and the evaluation metric is pixel-level F1 score only for manipulated images.
Based on the results, it is evident that our MSCCNet outperforms other models in terms of the average F1 score. 
Specifically, the localization results on CASIAv1 \cite{casiav1} far exceed other traditional image manipulation localization methods, indicating that our model is not only applicable to advanced face manipulation techniques but can also cover traditional manual image editing techniques.

\subsection{Ablation Study} \label{sec:experiment_3}
We analyze different modules of the proposed MSCCNet on the C40 set of P-FF++ and adopt the intra-dataset evaluation protocol.

%--------------Figure Visualization---------------
\begin{figure*}[!t]
\centering
\includegraphics[width=0.99\textwidth]{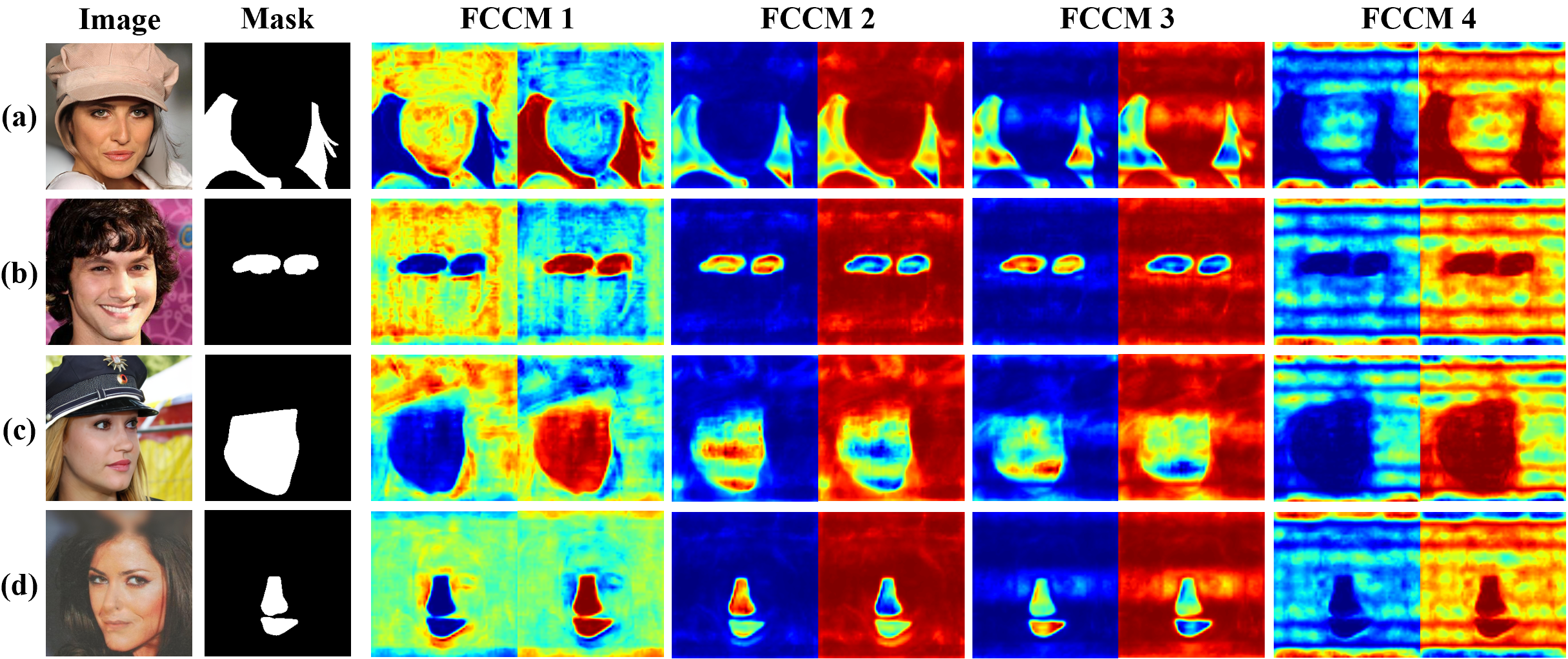}
\caption{Visualization of different frequency class center maps (FCCM). The Image column shows different types of manipulation methods, where (a), (b), (c), and (d) correspond to Repaint \cite{lugmayr2022RePaint}, LaMa \cite{rombach2022high}, LDM \cite{suvorov2021resolution}, and Pluralistic \cite{zheng2019pluralistic}, respectively. 
The Mask column indicates the tampered regions of the forged images.
The FCCM 1-4 columns represent the class center feature maps of the four frequency bands, indicating that our method effectively distinguishes the tampered regions from the authentic regions.
}
\label{fig:frequency_map}
\end{figure*}
%--------------Figure Visualization---------------

%--------------Table ablation-modules---------------
\begin{table}[!t]
\begin{center}
\caption{Analysis of different modules of the proposed MSCCNet.}
\label{tab:ab_modules}
\resizebox{0.99\linewidth}{!}{
\begin{tabular}{cccccccc}
\toprule
\multirow{2}{*}{Base.} & \multirow{2}{*}{MFA} & \multirow{2}{*}{MSCC} & \multirow{2}{*}{$\mathcal{L}_{mscc}$} & \multicolumn{2}{c}{Image-level} & \multicolumn{2}{c}{Pixel-level} \\ \cline{5-8} 
 &  &  &  & ACC & AUC & F1 & mIoU \\ \hline
\checkmark & - & - & - & 87.49 & 86.67 & 83.79 & 72.84 \\
\checkmark & \checkmark & - & - & 87.29 & 86.68 & 83.98 & 73.11 \\
\checkmark & \checkmark & \checkmark & - & 87.38 & 86.99 & 85.82 & 75.76 \\
\checkmark & \checkmark & \checkmark & \checkmark & 88.07 & 87.61 & 86.82 & 77.22 \\ 
\bottomrule
\end{tabular}
}
\end{center}
\end{table}
%--------------Table ablation-modules---------------

%--------------Table ablation-mscc---------------
\begin{table}[!t]
\begin{center}
\caption{Analysis of the proposed MSCC module.}
\label{tab:ab_mscc}
\resizebox{0.99\linewidth}{!}{
\begin{tabular}{cccccccc}
\toprule
\multirow{2}{*}{GCN} & \multirow{2}{*}{DCT} & \multirow{2}{*}{Add} & \multirow{2}{*}{Concat} & \multicolumn{2}{c}{Image-level} & \multicolumn{2}{c}{Pixel-level} \\ \cline{5-8} 
 &  &  &  & ACC & AUC & F1 & mIoU \\ \hline
\checkmark & \checkmark & - & \checkmark & 88.07 & 87.61 & 86.82 & 77.22 \\
- & \checkmark & - & \checkmark & 87.60 & 87.17 & 86.41 & 76.62 \\
\checkmark & - & - & \checkmark & 87.61 & 87.43 & 85.94 & 75.92 \\
\checkmark & \checkmark & \checkmark & - & 88.03 & 87.10 & 86.44 & 76.66 \\
% \checkmark & \checkmark & \checkmark & \checkmark& - & - & - & - \\
\bottomrule
\end{tabular}
}
\end{center}
\end{table}
%--------------Table ablation-mscc---------------

\subsubsection{Analysis on MSCCNet Architecture}
We set the baseline (\textit{Base.}) model by removing the MFA and MSCC modules, and the remaining convolutional blocks.
As summarized in Table \ref{tab:ab_modules}, applying the MFA module could bring $0.27\%$ mIoU improvements, which demonstrates that low-level local textures are helpful for manipulated region localization.
MSCC module is the key component for modeling semantic-agnostic features, it achieves $75.76\%$ in terms of mIoU.
The multi-spectral features of the coarse segmentation supervision mechanism enable the assessment of the probability of pixel attribution to its specific class. These features subsequently drive the MSCC module's ability to approximate a robust class center.
From the last line in Table \ref{tab:ab_modules}, we can observe that $\mathcal{L}_{mscc}$ improves the localization performance from $75.76\%$ to $77.22\%$. 
Our results show that the combination of semantic-agnostic features and low-level artifacts improves face manipulation localization. 
Moreover, the proposed MSCC module offers a viable solution to suppress semantic-related information through a multi-frequency perspective.
As shown in Fig. \ref{fig:frequency_map}, the visualization results of the multi-spectral class centers demonstrate that our MSCC module can learn distinct class centers for tampered and authentic regions, and effectively distinguish between them.

\subsubsection{Influence of GCN}
The GCN layer in our MSCC module improves the consistency of multi-spectral class-level representations by enhancing interaction between class centers across various frequency bands.
As can be seen in Table \ref{tab:ab_mscc}, if the GCN layer is removed, the localization performance drops from $77.22\%$ to $76.62\%$ mIoU.
It helps with multi-frequency attention map calculation in feature refinement operations.

%--------------Table ablation- frequency module---------------
\begin{table}[!t]
\begin{center}
\caption{Analysis of different frequency modules.}
\label{tab:ab_fre}
\resizebox{0.99\linewidth}{!}{
\begin{tabular}{ccccc}
\toprule
\multirow{2}{*}{\begin{tabular}[c]{@{}c@{}}Frequency Modules\end{tabular}} & \multicolumn{2}{c}{Image-level} & \multicolumn{2}{c}{Pixel-level} \\ \cline{2-5} 
 & ACC & AUC & F1 & mIoU \\ \hline
M2TR \cite{wang2021m2tr} & 85.09 & 86.13 & 83.65 & 72.61 \\
F3Net \cite{qian2020thinking} & 85.76 & 86.28 & 82.78 & 71.35 \\
Ours & 88.07 & 87.61 & 86.82 & 77.22 \\
\bottomrule
\end{tabular}
}
\end{center}
\end{table}
%--------------Table ablation-frequency module---------------

\subsubsection{Influence of DCT Filters}
In Sec. \ref{sec:method_3}, the DCT filters decompose semantic context features to different frequency bands, which relieves the aliasing among low-frequency and high-frequency components \cite{zhang2019making}.
Given that forgery traces are more prominent in high-frequency rather than low-frequency components \cite{qian2020thinking,liu2021spatial,chen2021local,luo2021generalizing,wang2021m2tr}, the multi-spectral class centers have the potential to model frequency-dependent forgery traces, particularly in high-frequency regions.
To show the effectiveness, we remove the DCT filters of the MSCC module, the performance drops to $75.92\%$.
In comparison, applying DCT filters brings $1.3\%$ mIoU improvements, as indicated in Table \ref{tab:ab_mscc}.

Furthermore, we compare the proposed MSCC module with the frequency-related module in face forgery detection methods (i.e., M2TR \cite{wang2021m2tr} and F3Net \cite{qian2020thinking}), as shown in Table \ref{tab:ab_fre}. 
Ours significantly outperforms them in terms of localization, which is because they are designed for detection tasks, while our MSCC module is specifically designed for the localization task.

\subsubsection{Influence of Fusion Type}
There are two feature fusion types: addition (Add) and concatenation (Concat) options for Eq. (\ref{eq16}).
In Table \ref{tab:ab_mscc}, we try both addition and concatenation, and the experimental results demonstrate that the concatenation type is better performance.

%--------------Table ablation-transformation basis---------------
\begin{table}[!t]
\begin{center}
\caption{Analysis of the number of the frequency spectrum.}
\label{tab:ab_bands}
\resizebox{0.99\linewidth}{!}{
\begin{tabular}{ccccc}
\toprule
\multirow{2}{*}{\begin{tabular}[c]{@{}c@{}}Number of \\ Frequency Spectrum\end{tabular}} & \multicolumn{2}{c}{Image-level} & \multicolumn{2}{c}{Pixel-level} \\ \cline{2-5} 
 & ACC & AUC & F1 & mIoU \\ \hline
1 & 87.68 & 86.83 & 86.23 & 76.34 \\
4 & 88.07 & 87.61 & 86.82 & 77.22 \\
16 & 88.06 & 87.44 & 86.46 & 76.70 \\
\bottomrule
\end{tabular}
}
\end{center}
\end{table}
%--------------Table ablation-transformation basis---------------

\subsubsection{Influence of the Number of Frequency Spectrum}
The number of the frequency spectrum of frequency maps can be denoted as $M$ in Sec. \ref{sec:method_3}.
Various experiments are conducted to investigate the performance of using different $M$, and the experimental results are shown in Table \ref{tab:ab_bands}.
Setting $M$ to $1$ indicates that the frequency features are not decomposed.
Thus, its mIoU is $0.88\%$ lower than $M=4$. 
Note that $M=4$ means the more frequency components are decomposed including low- and high-frequency.
We also notice that performance drops to $76.70$ if we use $M=16$.
This is primarily due to the increased difficulty of predicting accurate coarse segmentation outcomes for multi-frequency features, resulting in inadequate class-level representations when $M$ is too large.
Therefore, we adopt $M=4$ for the other experiments.

%-------------------------------------------------------------------------
\section{Conclusion}
This paper proposes a novel Multi-Spectral Class Center Network (MSCCNet) to facilitate learning more generalizable and frequency-related forgery features for improved face manipulation localization. 
To avoid reliance on semantic objective information, we employ a Multi-Spectral Class Center (MSCC) module to compute different frequency class-level contexts and weighted attention, enabling the refinement of forgery features in the localization network. 
A Multi-Level Feature Aggregation (MFA) module is integrated to fuse low-level forgery-specific textures. 
We construct a localization performance benchmark consisting of deepfake detection and image manipulation localization methods, based on the reconstructed P-FF++ dataset and the latest diffusion-based Dolos dataset. 
Our extensive experiments demonstrate the superior localization ability of MSCCNet on the comprehensive benchmarks introduced.

\bibliographystyle{IEEEtran}
\bibliography{bibs/cvpr,bibs/tcsvt}

\end{document}